\documentclass[letterpaper, 10 pt, conference]{ieeeconf}  

\IEEEoverridecommandlockouts{}                              

\usepackage{amsmath,amsfonts}
\usepackage{algorithmic}
\usepackage{algorithm}
\usepackage{array}
\usepackage{textcomp}
\usepackage{stfloats}
\usepackage{url}
\usepackage{verbatim}
\usepackage{graphicx}
\usepackage{cite}
\usepackage{booktabs}
\usepackage{pgfplots}
\usepackage{capt-of} 
\pgfplotsset{compat=1.18}
\IEEEoverridecommandlockouts
\title{\LARGE \bf
ReTac-ACT: A State-Gated Vision-Tactile Fusion Transformer for Precision Assembly
}

\author{
\authorblockN{Minchi Ruan$^{1,2,\dagger}$, LiangQing Zhou$^{1,2,\dagger}$, Hongtong Li$^{1}$, Zongtao Wang$^{2}$, ZhaoMing Lu$^{1}$, Jianwei Zhang$^{3}$, Bin Fang$^{1,2,\ast}$}
\authorblockA{
$^{1}$Beijing University of Posts and Telecommunications, Beijing, China\\
$^{2}$SunHDex Intelligent Technology (Beijing) Co., Ltd., Beijing, China\\
$^{3}$University of Hamburg, Hamburg, Germany\\
$^{\dagger}$These authors contributed equally.\\
$^{\ast}$Corresponding author
}
}

\begin{document}

\twocolumn[{%
\renewcommand\twocolumn[1][]{#1}%
\maketitle
\begin{center}
  \includegraphics[width=0.98\textwidth]{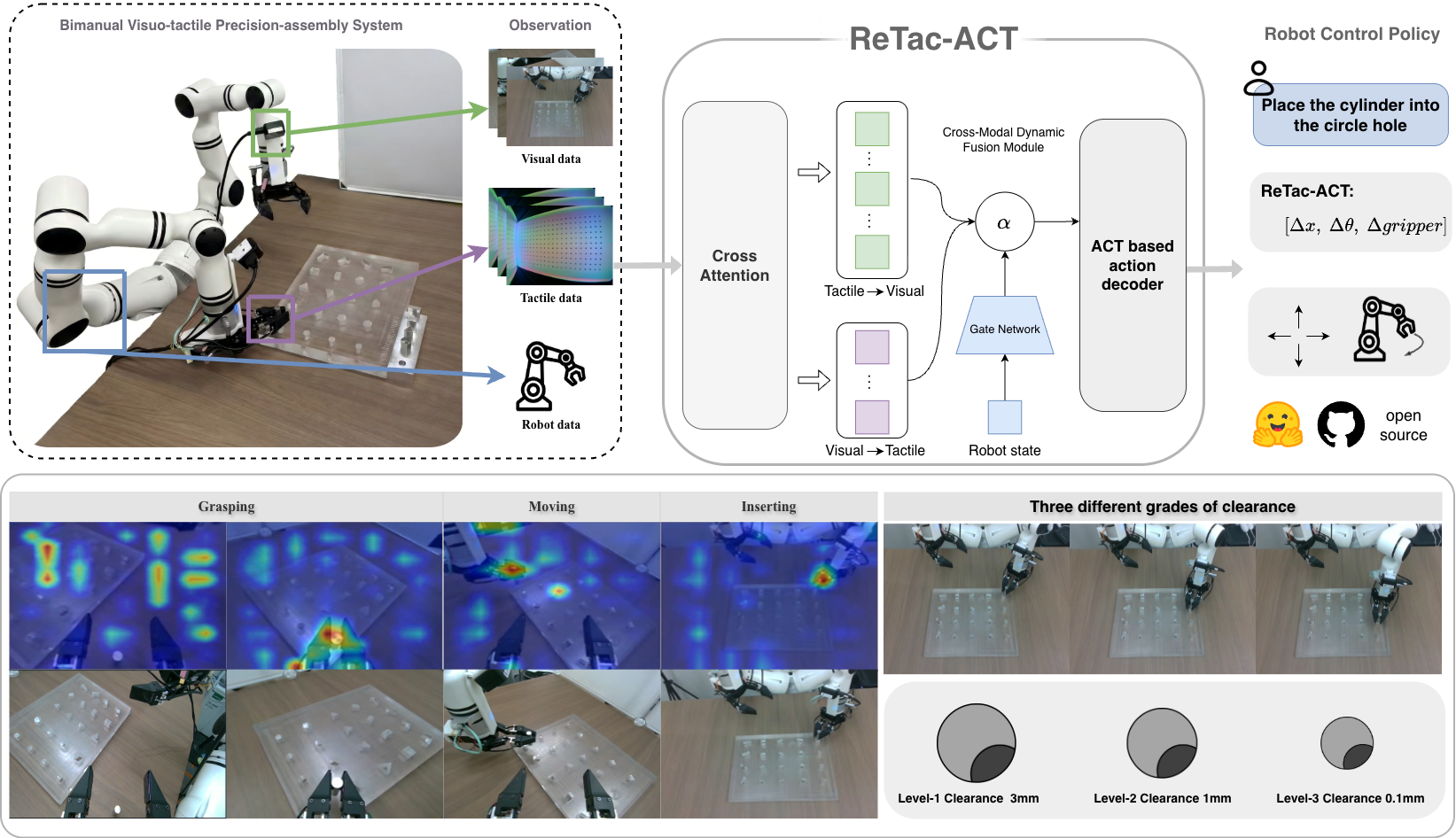}
  \captionof{figure}{We present ReTac-ACT, a state-gated vision-tactile policy that extends Action Chunking with Transformers (ACT) to natively process tactile feedback. ReTac-ACT sets a new state of the art for high-precision peg-in-hole tasks on the NIST ATB M1 benchmark provided by ManipulationNet, achieving 90\% success at 3\,mm clearance and maintaining 80\% at industrial-grade 0.1\,mm clearance where pure vision fails due to occlusion. It features a proprioception-conditioned gating mechanism to dynamically fuse visual and tactile modalities and is trained with auxiliary tactile reconstruction objectives. The ReTac-ACT code will be made open-source to support the research community.}
  \label{fig:system_overview}
\end{center}%
}]
\thispagestyle{empty}
\pagestyle{empty}

\begin{abstract}

Precision assembly requires sub-millimeter corrections in contact-rich ``last-millimeter'' regions where visual feedback fails due to occlusion from the end-effector and workpiece.
We present ReTac-ACT (Reconstruction-enhanced Tactile ACT), a vision-tactile imitation learning policy that addresses this challenge through three synergistic mechanisms:
(i) bidirectional cross-attention enabling reciprocal visuo-tactile feature enhancement before fusion,
(ii) a proprioception-conditioned gating network that dynamically elevates tactile reliance when visual occlusion occurs, and
(iii) a tactile reconstruction objective enforcing learning of manipulation-relevant contact information rather than generic visual textures.
Evaluated on the standardized NIST Assembly Task Board M1 benchmark, ReTac-ACT achieves 90\% peg-in-hole success, substantially outperforming vision-only and generalist baseline methods, and maintains 80\% success at industrial-grade 0.1\,mm clearance.
Ablation studies validate that each architectural component is indispensable.
The ReTac-ACT codebase and a vision-tactile demonstration dataset covering various clearance levels with both visual and tactile features will be released to support reproducible research.

\end{abstract}


\section{Introduction}
In human manufacturing, the synergistic integration of vision and touch constitutes a core capability for performing precision operations. Consider high-precision peg-in-hole assembly as an illustrative example: visual perception is responsible for the semantic recognition and coarse alignment of components, whereas tactile feedback provides critical real-time force information during the contact phase, guiding dynamic millimeter-level adjustments. This multimodal perception paradigm significantly enhances operational robustness, particularly in contact-intensive scenarios where visual information is limited, occlusions are severe, or geometric features are indistinct. In such contexts, tactile signals frequently serve as the decisive determinant of task success.

In recent years, vision-based methods have demonstrated excellent performance in various robotic manipulation tasks. Representative works, such as Diffusion Policy~\cite{ref7}, ACT~\cite{ref9}, and pi05~\cite{ref1}, have achieved high success rates in general-purpose manipulation tasks. However, vision-only Imitation Learning (IL) methods~\cite{ref1,ref2,ref7,ref9,ref16} continue to face significant challenges in precision operations, including millimeter-level peg-in-hole assembly. The fundamental limitation of these methods lies in their over-reliance on visual perception, which neglects crucial tactile feedback during the contact phase. This dependency renders the system highly susceptible to failure under conditions of visual occlusion and geometric ambiguity. Fortunately, with the rapid development of high-precision tactile sensor technology~\cite{ref8,ref27,ref28,ref29,ref30}, rich and reliable contact information can now be integrated effectively into robot control loops, establishing the hardware foundation for genuine visuo-tactile fusion.

To address these challenges, we propose ReTac-ACT, a novel vision-tactile policy designed to enable Action Chunking with Transformers (ACT) to process and utilize tactile feedback natively. While the original ACT framework excels at vision-based trajectory tracking, it lacks the mechanisms required to incorporate the high-dimensional tactile signals essential for contact-rich tasks. ReTac-ACT bridges this gap by introducing a state-gated dynamic fusion module that adaptively weights visual and tactile modalities based on the proprioceptive state of the robot. This design enables the policy to transition seamlessly from a vision-dominant mode during the free-space approach to a tactile-dominant mode during precision insertion, thereby robustly handling visual occlusion and complex contact dynamics.

The evaluation of precision assembly policies has historically been hindered by the use of custom, non-standardized setups where component dimensions, machining tolerances, and clearance levels are rarely disclosed. To ensure strict reproducibility and meaningful performance comparisons, we rigorously evaluate our method on the open-source NIST Assembly Task Board (ATB) M1 benchmark provided by ManipulationNet~\cite{manipulationnet2025}. In this benchmark, the exact machining tolerances of the pegs and the clearance levels of the task board are explicitly standardized and publicly available.

Our main contributions are summarized as follows:
\begin{itemize}
  \item We propose ReTac-ACT, a state-gated vision-tactile Action Chunking Transformer that extends the ACT architecture to be natively compatible with tactile inputs. The framework features separate encoders for each modality and a gating network conditioned on proprioception that dynamically fuses visual and tactile feature streams, ensuring robust policy execution across different task phases.
  \item We introduce a tactile representation learning scheme utilizing an auxiliary reconstruction objective during training. This objective compels the tactile encoder to capture explicit high-frequency contact geometry instead of collapsing into trivial features, thereby significantly enhancing the sensitivity of the policy to sub-millimeter contact deviations.
  \item We release a vision-tactile peg-in-hole assembly dataset comprising over 5,000 expert demonstration trajectories, covering 5 geometric shapes and 4 assembly tolerance levels based on the NIST ATB M1 task board, together with the complete implementation codebase for ReTac-ACT.
\end{itemize}

\begin{figure*}[t]
  \centering
  \includegraphics[width=\linewidth]{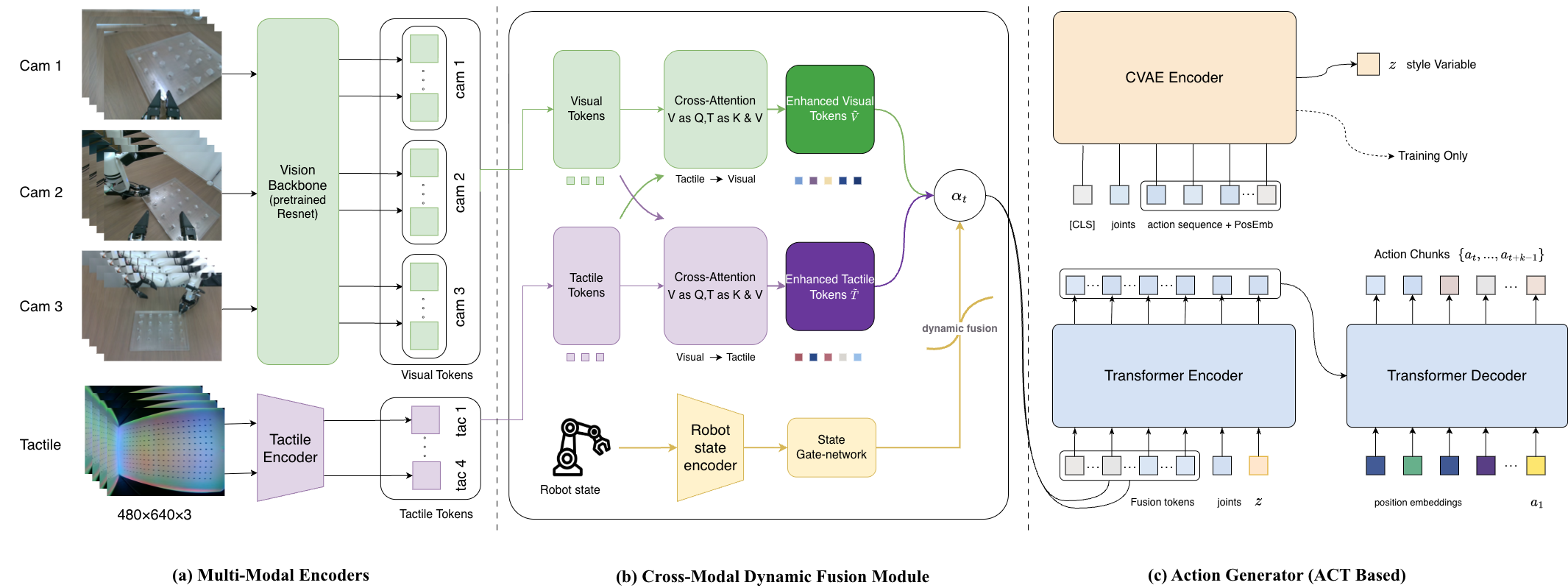}
  \caption{Overview of the ReTac-ACT architecture. (a) \textbf{Multi-Modal Encoders}: Visual inputs (3 RGB cameras) and tactile inputs (4 contact images: one sensor per fingertip, two fingers per gripper, bimanual) are processed by separate backbones into feature tokens. (b) \textbf{Cross-Modal Dynamic Fusion}: A proprioception-based gating network dynamically weighs modalities, enhanced by bidirectional cross-attention. (c) \textbf{Action Generator}: A CVAE-based transformer decoder predicts temporal action chunks $\hat{a}_{t:t+k-1}$, where each action includes 14-DoF bimanual joint targets and 2 gripper commands.}
  \label{fig:model_arch}
\end{figure*}

\section{Related Work}

\textbf{Vision-Tactile Precision Assembly:} Tactile perception is crucial for robots to perform precision assembly tasks. In scenarios with visual occlusion or a lack of sufficient visual features, tactile feedback can provide key physical contact signals, thereby ensuring the smooth execution of precision operations. Early explorations~\cite{ref31, ref32} mainly relied on RGB cameras and force/torque (F/T) sensors to infer contact states and make decisions. However, the low-dimensional information provided by F/T sensors is insufficient to meet the demands of complex and dexterous manipulation. As a result, tactile sensors capable of providing high-dimensional features have been widely adopted in robotic manipulation. In actual industrial production environments, traditional capacitive or pressure-sensitive tactile sensors are highly susceptible to electromagnetic or physical interference, leading to signal distortion. To address this issue in our study, when the robot arm's field of view (FOV) near the contact point is occluded---rendering purely vision-based pose estimation unreliable---we utilize optical tactile sensors such as GelSight~\cite{ref8, ref19} to detect multi-directional contact deformations. Achieving robust assembly insertion necessitates the full utilization of such high-resolution and rich contact cues.

\textbf{Vision-Tactile Learning and Fusion:} In recent years, learning-based methods have demonstrated strong performance in integrating vision, proprioception, and tactile perception, particularly in contact-rich manipulation tasks. In recent studies, Calandra et al.~\cite{ref20} utilized deep neural networks, and Li et al.~\cite{ref15} used the self-attention mechanism in Transformers to extract and process visual and tactile information. With the significant advancements of Vision-Language Models (VLMs) in the multimodal domain, methods employing VLMs to fuse visual and tactile features (e.g., TLA~\cite{ref5, ref21} and VTLA~\cite{ref11,ref17,ref25,ref26}) are increasingly prevalent. Most of these approaches adopt early fusion with parallel multimodal inputs. However, since tactile data is often weaker than visual data in terms of scale and information density, parallel input architectures can easily lead to ``modality imbalance.'' In this state, the model is easily dominated by visual signals, thereby diminishing the contribution of tactile information during fine manipulation. Existing studies have attempted to enhance the representational capacity of tactile features by introducing mechanisms such as masking and reconstruction~\cite{ref14,ref22,ref23,ref24}. Nevertheless, precision assembly typically encompasses two distinct stages: ``approach/grasping'' and ``insertion/assembly.'' During free-space motion or the post-grasping hold stage, tactile signals remain largely static. At this juncture, blindly amplifying the proportion of tactile features is not only counterproductive but also leads to redundant computational overhead during training and inference. To overcome these challenges, this paper not only adopts a tactile reconstruction mechanism to enhance tactile representation but also introduces a state-gated dynamic fusion mechanism. This mechanism enables the policy to adaptively switch between ``vision-dominated free-space motion'' and ``tactile-dominated contact manipulation,'' which significantly enhances the interpretability and robustness of the policy while improving the overall assembly success rate.

\textbf{Robotic Imitation Learning:} Imitation Learning (IL)~\cite{ref10} provides an effective paradigm for robots to efficiently acquire complex manipulation skills from expert demonstration data. Recent research has significantly advanced IL across policy architectures, computing platforms, and data processing paradigms. For instance, Action Chunking with Transformers (ACT) proposed by Zhao et al.~\cite{ref9} constructs a powerful behavioral cloning policy. By predicting chunks of actions, this algorithm exhibits exceptional sequence modeling capabilities and achieves temporally extended action generation. Building upon this foundation, subsequent research has yielded improved architectures such as Bi-LAT~\cite{ref4} for dual-arm collaboration and methods that integrate Diffusion Policy~\cite{ref6}, as well as RTC~\cite{ref13}. However, these methods heavily rely on visual inputs and struggle to reliably execute fine manipulation tasks under complex working conditions, such as severe visual occlusion. To bridge this gap, this paper introduces high-dimensional tactile inputs into the ACT framework and further extends it with a state-gated bidirectional vision-tactile fusion mechanism, effectively tackling high-precision assembly challenges in complex, occluded environments.

\section{Method}
\subsection{Overview}

\subsubsection{Notation}
Let $x_t$ denote the robot proprioceptive state at time $t$, $I^v_t$ the visual observations, and $I^t_t$ the tactile observations.
We encode these into visual tokens $V\in\mathbb{R}^{N_v\times D}$ and tactile tokens $T\in\mathbb{R}^{N_t\times D}$, where $D$ is the feature dimension.
The policy outputs an action chunk $\{a_t,\ldots,a_{t+k-1}\}$ of length $k$.
We use $\tilde{V},\tilde{T}$ for cross-attention-enhanced features, and $V^\star, T^\star$ for the final fused representations.

\subsubsection{Architecture Overview}
As illustrated in Fig.~\ref{fig:model_arch}, ReTac-ACT comprises three modules: (1) modality-specific encoders for visual and tactile inputs, (2) a state-gated cross-modal fusion module conditioned on robot proprioception, and (3) a CVAE-based transformer decoder for action chunk prediction.
We adopt the ACT framework~\cite{ref9} for action generation: a Conditional VAE (CVAE) encodes the ground-truth action chunk into a latent variable $z$ during training.
A 4-layer Transformer Encoder processes the fused tokens $[V^{\star}; T^{\star}]$ concatenated with the proprioceptive embedding.
A 1-layer Transformer Decoder then predicts the action chunk $\{a_t,\ldots,a_{t+k-1}\}$ conditioned on the encoder output and latent $z$.
During inference, $z$ is sampled from the prior mean.

\subsection{Multi-Modal Encoders}
We separately encode visual and tactile modalities before fusion to allow specialization in each sensory domain.

\begin{figure}[t]
  \centering
  \includegraphics[width=\linewidth]{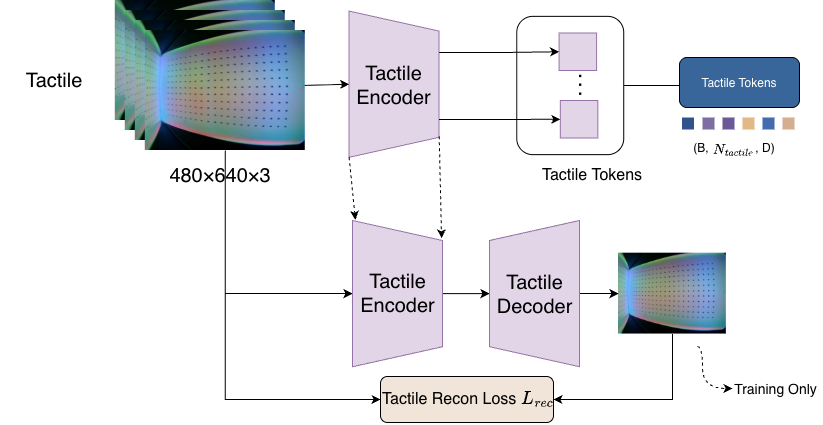}
  \caption{Tactile representation learning via auxiliary reconstruction. During training, an image reconstruction objective regularizes the tactile encoder. By employing a decoder to reconstruct the raw tactile inputs from the learned latent tokens, the model is explicitly forced to capture fine-grained contact geometry, preventing feature collapse.}
  \label{fig:tactile_recon}
\end{figure}

\emph{1) Visual Encoder:} For each RGB camera, we extract multi-scale features via a ResNet-18 backbone (pretrained on ImageNet), producing feature maps $F^v\in\mathbb{R}^{C_v\times H_v\times W_v}$.
We spatially flatten and linearly project each map to obtain per-camera visual tokens $V_i\in\mathbb{R}^{N_v\times D}$.
The three camera streams are concatenated to form the full visual token set $V\in\mathbb{R}^{3N_v\times D}$.

\emph{2) Tactile Encoder:} Directly employing standard visual backbones for tactile images is suboptimal. Models pre-trained on natural images (e.g., ImageNet) prioritize semantic features and struggle to capture the high-frequency contact deformations essential for insertion due to the significant domain gap. To address this, we employ a dedicated 5-layer CNN encoder with progressively increasing channels \{32,64,128,256,512\}, kernel size 4, and stride 2.
The resulting feature map is flattened and projected to $T\in\mathbb{R}^{N_t\times D}$ ($N_t{=}196$ tokens). As depicted in Fig.~\ref{fig:tactile_recon}, this encoder is regularized during training via an auxiliary tactile reconstruction objective. By employing a decoder to reconstruct the raw tactile inputs from the learned latent tokens, we explicitly force the network to preserve fine-grained contact geometry and prevent feature collapse, which is vital for high-precision insertion.

\subsection{Cross-Modal Dynamic Fusion Module}
Precision assembly inherently involves drastic shifts in modality reliability: vision dominates during the free-space approach, whereas tactile feedback becomes critical during occluded insertion. To seamlessly handle these phase transitions without requiring explicit heuristic switches, we propose a Cross-Modal Dynamic Fusion Module comprising two key stages. First, a bidirectional cross-attention mechanism enables reciprocal feature enhancement between visual and tactile tokens. Second, a proprioception-conditioned gating mechanism dynamically adjusts the fusion weights, ensuring the policy adaptively relies on the most informative modality.

\textbf{Bidirectional Cross-Modal Attention:} Before fusion, we enable mutual enhancement between modalities via cross-attention.
Given visual tokens $V$ and tactile tokens $T$, we compute:
\begin{equation}
  \begin{aligned}
    \tilde{T} &= \mathrm{LN}\big(T + \mathrm{MHA}_{\mathrm{T\to V}}(Q{=}T,\,K{=}V,\,V{=}V)\big), \\
    \tilde{V} &= \mathrm{LN}\big(V + \mathrm{MHA}_{\mathrm{V\to T}}(Q{=}V,\,K{=}T,\,V{=}T)\big),
  \end{aligned}
\end{equation}
where $\tilde{T}$ represents visual-context-enriched tactile features, and $\tilde{V}$ represents tactile-refined visual features.

\textbf{Gating Mechanism:} We compute a scalar modality gate $\alpha_t\in(0,1)$ from the proprioceptive state $x_t$ via a learned 3-layer MLP $g(\cdot)$:
\begin{equation}
  \alpha_t = \sigma\big(g(x_t)/\tau_g\big),
\end{equation}
where $\sigma$ is the sigmoid function and $\tau_g>0$ is a learnable temperature parameter (initialized to 1.0, clamped at minimum 0.1) that controls the sharpness of modality switching.
When $\alpha_t\approx 0$, the policy favors vision (free-space approach); when $\alpha_t\approx 1$, it emphasizes tactile cues (contact phase).

\textbf{Reciprocal Fusion:} We fuse the cross-attention-enhanced features in a reciprocal manner:
\begin{equation}
  \begin{aligned}
    V^\star &= (1-\alpha_t)\,V + \alpha_t\,\tilde{V}, \\
    T^\star &= \alpha_t\,T + (1-\alpha_t)\,\tilde{T}.
  \end{aligned}
\end{equation}
This design creates an information bottleneck: increasing reliance on one modality proportionally down-weights the other, preventing the policy from trivially concatenating all signals and forcing meaningful modality-switching behavior.
The final fused representation $[V^\star; T^\star]$ is then passed to the transformer encoder.

\subsection{Loss Function}

Beyond action prediction loss $\mathcal{L}_{\ell_1}$ and VAE KL-divergence $\mathcal{L}_{\mathrm{KL}}$, we introduce auxiliary objectives to enforce structured representation learning.

\textbf{Tactile Reconstruction:}
Standard visual backbones tend to treat tactile images as low-level textures, losing the contact geometry critical for precision assembly. To enforce learning of manipulation-relevant tactile representations, we employ an auxiliary reconstruction objective during training.
A transposed-convolution decoder (mirroring the encoder architecture) reconstructs the input tactile image $\hat{I}^t_t$ from the encoded tactile tokens.
The reconstruction loss is:
\begin{equation}
  \mathcal{L}_{\mathrm{rec}} = \|\hat{I}^t_t - I^t_t\|_2^2.
\end{equation}
This self-supervised signal forces the encoder to preserve fine-grained contact deformation patterns essential for sub-millimeter insertion corrections, preventing feature collapse particularly during free-space motion where tactile signals remain static.

\textbf{Contrastive Alignment:}
To ensure semantic alignment between visual and tactile feature spaces despite their domain gap, we employ an InfoNCE-style contrastive loss. We project the globally pooled visual and tactile tokens into a shared latent space and maximize similarity between synchronized vision-tactile pairs while minimizing it for mismatched pairs:
\begin{equation}
  \mathcal{L}_{\mathrm{con}} = -\log \frac{\exp(\mathrm{sim}(e^v_i, e^t_i)/\tau_{\mathrm{con}})}{\sum_{j=1}^{B} \exp(\mathrm{sim}(e^v_i, e^t_j)/\tau_{\mathrm{con}})},
\end{equation}
where $e^v_i, e^t_i$ denote the projected visual and tactile embeddings for the $i$-th sample, $B$ is the batch size, and $\tau_{\mathrm{con}}$ is a temperature hyperparameter.

\textbf{Joint Optimization:}
The total objective combines all losses:
\begin{equation}
  \mathcal{L} = \mathcal{L}_{\ell_1} + \lambda_{\mathrm{KL}}\mathcal{L}_{\mathrm{KL}} + \lambda_{\mathrm{rec}}\mathcal{L}_{\mathrm{rec}} + \lambda_{\mathrm{con}}\mathcal{L}_{\mathrm{con}}.
\end{equation}
We set weights $\lambda_{\mathrm{KL}}{=}10$, $\lambda_{\mathrm{rec}}{=}0.5$, $\lambda_{\mathrm{con}}{=}0.1$. The model is trained end-to-end using AdamW with learning rate $10^{-5}$ for the backbone and $10^{-4}$ for the tactile encoder, weight decay $10^{-4}$, action chunk size $k{=}100$, model dimension $D{=}512$, 8 attention heads, and latent dimension 32.

\section{Experiments}

\subsection{Experimental Setup}

\subsubsection{Task and Hardware}
\begin{figure}[t]
  \centering
  \includegraphics[width=\linewidth]{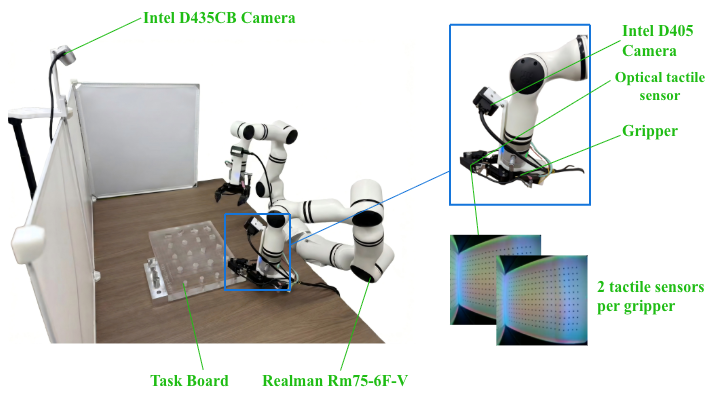}
  \caption{Hardware setup for the bimanual vision-tactile precision-assembly system.}
  \label{fig:hardware}
\end{figure}

We evaluate on the NIST ATB M1 precision assembly benchmark, which requires inserting cylindrical pegs into corresponding holes at three clearance levels: Level~1 (3\,mm), Level~2 (1\,mm), and Level~3 (0.1\,mm).
All pegs are 304 stainless steel with CNC tolerance below 10$\,\mu\mathrm{m}$.
The hardware setup is shown in Fig.~\ref{fig:hardware}. It comprises a bimanual robot with two Realman RM75-6F-V arms (7-DOF each, 14-DOF total), each equipped with a parallel gripper. Proprioception ($x_t$) includes joint positions, Cartesian poses, and gripper states. Visual observations ($I^v_t$) are captured by multi-view RGB cameras: two wrist-mounted plus one third-person view at $480{\times}640$. Tactile observations ($I^t_t$) are provided by four Xense optical tactile sensors (two per gripper), each capturing high-resolution contact images at $480{\times}640$.

\subsubsection{Data Collection}
To build a comprehensive resource for the community, we collected a large-scale vision-tactile dataset comprising over 5,000 expert demonstration trajectories across 5 geometric shapes and 4 clearance levels. The data was collected via teleoperation using VR controllers. Each trajectory records synchronized multimodal data at a sampling frequency of 30\,Hz, including three-view visual images (stored as videos), bimanual joint angles, end-effector Cartesian poses (position and quaternion orientation relative to the robot base frame), high-resolution tactile images from the Xense optical tactile sensors, and the corresponding action commands. All data modalities are strictly time-synchronized and standardized into the LeRobot dataset V3.0 format. To ensure dataset diversity and robustness, we randomized the initial poses of the robot and the peg for each episode, and data was collected by multiple human operators to prevent the policy from overfitting to a single operator's behavioral habits.
While we provide this extensive dataset for future large-scale pre-training research, our proposed ReTac-ACT is highly sample-efficient. For the experiments reported in this paper, training a successful policy for a specific clearance level required only 50 to 100 demonstration trajectories.

\subsubsection{Evaluation Protocol}
We report three metrics capturing the full manipulation pipeline: (1) \textbf{Missed}: percentage of trials where grasp fails entirely; (2) \textbf{Grasp}: percentage of trials with successful grasp; (3) \textbf{Peg-in-hole}: final insertion success rate.

\subsubsection{Baselines}
We compare against two widely adopted BC methods and one state-of-the-art open-source generalist VLA policy:
(i) \textbf{ACT}~\cite{ref9}: predicts action chunks via a CVAE-based transformer with temporal ensembling, targeting fine-grained manipulation. As the direct architectural predecessor of ReTac-ACT, this comparison isolates the contribution of our vision-tactile extensions.
(ii) \textbf{Diffusion Policy (DP)}~\cite{ref7}: models visuomotor policy as a conditional denoising diffusion process, generating actions by iteratively denoising Gaussian noise conditioned on visual observations. It naturally handles multimodal action distributions and is widely adopted for its training stability.
(iii) \textbf{pi05}~\cite{ref1}: the flow-matching variant of $\pi_0$, a generalist VLA model built on a pre-trained VLM backbone and trained on large-scale multi-robot data. This baseline tests whether broad generalist pre-training can substitute for dedicated tactile sensing on our specialized high-precision task.

\begin{table}[htbp]
  \centering
  \caption{Quantitative results for peg-in-hole assembly at Level~1 (3\,mm clearance). The best results are highlighted in bold.}
  \label{tab:main_results_baselines}
  \begin{tabular}{lccc}
    \toprule
    Method & Missed & Grasp & Peg-in-hole \\
    \midrule
    ACT~\cite{ref9} & 40\% & 60\% & 40\% \\
    DP~\cite{ref7} & 70\% & 30\% & 20\% \\
    pi05~\cite{ref1} & 55\% & 45\% & 20\% \\
    \textbf{ReTac-ACT (ours)} & \textbf{0\%} & \textbf{100\%} & \textbf{90\%} \\
    \bottomrule
  \end{tabular}
\end{table}

\subsection{Comparison with Baseline Methods}
Tests are conducted with cylindrical pegs placed at two different initial positions, 10 trials per position (20 trials total). We compare ReTac-ACT against behavior cloning (ACT, Diffusion Policy) and vision-language-action (pi05) baselines.
ReTac-ACT achieves 90\% peg-in-hole success, outperforming all baselines by a significant margin (Table~\ref{tab:main_results_baselines}).

\emph{1) Grasping Phase:}
Baseline methods exhibit high grasp failure rates: ACT misses 40\%, DP misses 70\%, and pi05 misses 55\%.
In contrast, ReTac-ACT achieves 0\% miss rate and 100\% grasp success.
This improvement stems from our bidirectional cross-attention mechanism, which enables tactile features to enhance visual localization of the target object even before contact, while visual context reciprocally guides tactile-informed grasp refinement.

\emph{2) Insertion Phase:}
ReTac-ACT achieves 90\% insertion success versus 40\% for ACT, a 2.25-fold improvement.
DP and pi05 achieve only 20\%.
The state-gated dynamic fusion mechanism plays a critical role here: upon contact, the proprioception-conditioned gate shifts reliance from vision to tactile, enabling sub-millimeter corrections that vision alone cannot provide due to occlusion.

\emph{3) VLA Generalist Limitation:}
Despite being a state-of-the-art pre-trained model, pi05 achieves only 20\% success even with extended training.
This confirms that generalist VLA models, lacking tactile pre-training, struggle on high-precision contact-rich tasks, highlighting the necessity of specialized vision-tactile fusion.

\begin{table}[htbp]
  \centering
  \caption{Ablation study on key components of ReTac-ACT. The best results are highlighted in bold.}
  \label{tab:main_results_ablation}
  \begin{tabular}{lccc}
    \toprule
    Variant & Missed & Grasp & Peg-in-hole \\
    \midrule
    w/o Reciprocal Fusion & 0\% & 100\% & 5\% \\
    w/o Cross-Attention & 0\% & 100\% & 5\% \\
    w/o TacRecon & 5\% & 95\% & 15\% \\
    w/o StateGate & 0\% & 100\% & 35\% \\
    \textbf{ReTac-ACT (full)} & \textbf{0\%} & \textbf{100\%} & \textbf{90\%} \\
    \bottomrule
  \end{tabular}
\end{table}

\subsection{Ablation Study}
To validate the necessity of each architectural component, we conduct a systematic ablation study following the same evaluation protocol as Table~\ref{tab:main_results_baselines}.
Our full ReTac-ACT model achieves superior performance across all phases: 0\% miss rate, 100\% grasp success, and 90\% insertion success.
To isolate individual contributions, we evaluate four ablated variants by selectively removing one module at a time: (1) without Reciprocal Fusion, (2) without bidirectional Cross-Attention, (3) without TacRecon, and (4) without StateGate.
Each ablation maintains the remaining architecture intact to ensure fair comparison.

Table~\ref{tab:main_results_ablation} demonstrates that every component is indispensable: removing any single module results in substantial performance degradation.
This systematic degradation pattern confirms that our architecture achieves high performance not through redundant design but through synergistic integration of complementary mechanisms.
We now analyze the specific role of each component:

\paragraph{Cross-Attention and Reciprocal Fusion Are Critical for Contact-Rich Manipulation}
Removing either bidirectional cross-attention or reciprocal fusion drastically reduces insertion success from 90\% to 5\%.
This severe degradation demonstrates that effective vision-tactile integration requires explicit bidirectional information exchange beyond simple feature concatenation.
The cross-attention mechanism enables tactile features to enhance visual localization while visual context guides tactile interpretation, crucial when occlusion dominates during contact-rich insertion.

\paragraph{Tactile Reconstruction Provides Manipulation-Relevant Representations}
Removing the tactile reconstruction module (TacRecon) degrades performance across all phases: miss rate increases to 5\%, grasp success drops to 95\%, and insertion success falls to 15\%.
The reconstruction objective ($\mathcal{L}_{\mathrm{rec}}$) enforces that the tactile encoder captures contact geometry and deformation patterns rather than generic visual textures, providing features essential for sub-millimeter corrections.
The 15\% insertion success---better than the 5\% without cross-attention---confirms that even degraded tactile representations provide some benefit when properly fused, but fall far short of reconstruction-learned features.

\paragraph{State-Gated Fusion Enables Adaptive Modality Reweighting}
The state-gated dynamic fusion mechanism (StateGate) improves insertion success from 35\% to 90\% while maintaining 100\% grasp performance.
As formulated in Eq.~(3), the proprioception-conditioned gating network computes $\alpha_t$ to control fusion weights between visual and tactile tokens, enabling the policy to adaptively emphasize the more informative modality at each task phase.
The 35\% performance with static fusion (where $\alpha_t$ is fixed throughout execution) demonstrates that multimodal information is available but underutilized: the policy cannot sufficiently shift toward tactile dominance during the insertion phase when visual feedback degrades due to occlusion.
In contrast, the learned gating mechanism automatically increases tactile weighting upon contact, allowing the policy to exploit contact geometry for sub-millimeter corrections that vision alone cannot provide.
Notably, this 55 percentage point gain depends on the synergy with the previous two mechanisms: TacRecon provides discriminative tactile representations that convey manipulation-relevant geometry, and cross-attention enables bidirectional information exchange between modalities. Without high-quality tactile features or effective cross-modal interaction, adaptive gating has insufficient signal to leverage.

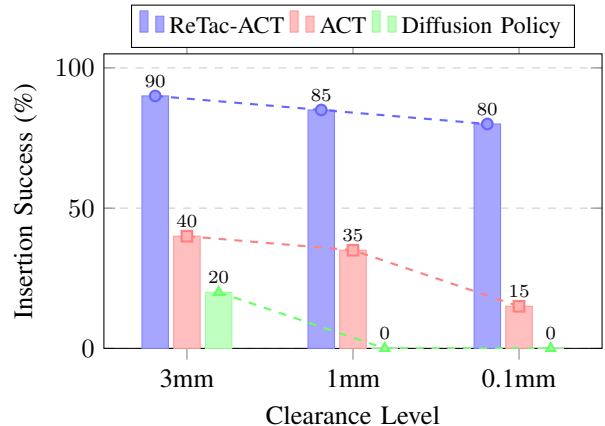
\begin{figure}[htbp]
  \centering
  \begin{tikzpicture}
    \begin{axis}[
      ybar,
      bar width=10pt,
      width=0.95\linewidth,
      height=5.5cm,
      ymin=0, ymax=105,
      ylabel={Insertion Success (\%)},
      xlabel={Clearance Level},
      symbolic x coords={3mm, 1mm, 0.1mm},
      xtick=data,
      legend style={at={(0.5,1.02)}, anchor=south, legend columns=3, font=\small},
      ymajorgrids=true,
      grid style={dashed, gray!40},
      nodes near coords,
      nodes near coords align={vertical},
      every node near coord/.append style={font=\scriptsize},
      enlarge x limits=0.25,
    ]
      \addplot[fill=blue!35, draw=blue!50, bar shift=-12pt] coordinates {(3mm,90) (1mm,85) (0.1mm,80)};
      \addplot[fill=red!25, draw=red!40, bar shift=0pt] coordinates {(3mm,40) (1mm,35) (0.1mm,15)};
      \addplot[fill=green!25, draw=green!40, bar shift=12pt] coordinates {(3mm,20) (1mm,0) (0.1mm,0)};
      \legend{ReTac-ACT, ACT, Diffusion Policy}
    \end{axis}
    \begin{axis}[
      width=0.95\linewidth,
      height=5.5cm,
      ymin=0, ymax=105,
      symbolic x coords={3mm, 1mm, 0.1mm},
      xtick=\empty,
      ytick=\empty,
      axis x line=none,
      axis y line=none,
      enlarge x limits=0.25,
    ]
      \addplot[thick, blue!60, dashed, mark=*, mark options={solid, fill=blue!40}, xshift=-12pt] coordinates {(3mm,90) (1mm,85) (0.1mm,80)};
      \addplot[thick, red!50, dashed, mark=square*, mark options={solid, fill=red!30}, xshift=0pt] coordinates {(3mm,40) (1mm,35) (0.1mm,15)};
      \addplot[thick, green!60, dashed, mark=triangle*, mark options={solid, fill=green!35}, xshift=12pt] coordinates {(3mm,20) (1mm,0) (0.1mm,0)};
    \end{axis}
  \end{tikzpicture}
  \caption{Robustness to tighter clearances. As clearance tightens from 3\,mm to 0.1\,mm, ReTac-ACT degrades only 11\% (90\%$\to$80\%) while ACT degrades 62.5\% (40\%$\to$15\%) and Diffusion Policy (DP) fails (20\%$\to$0\%). Dashed lines highlight the degradation trends.}
  \label{fig:clearance_comparison}
\end{figure}

\subsection{Robustness to Tighter Tolerances}
Industrial assembly often requires sub-millimeter precision.
To evaluate generalization to tighter tolerances, we compare ReTac-ACT against baselines across three clearance levels (Fig.~\ref{fig:clearance_comparison}).

\begin{figure*}[t]
  \centering
  \includegraphics[width=0.95\textwidth]{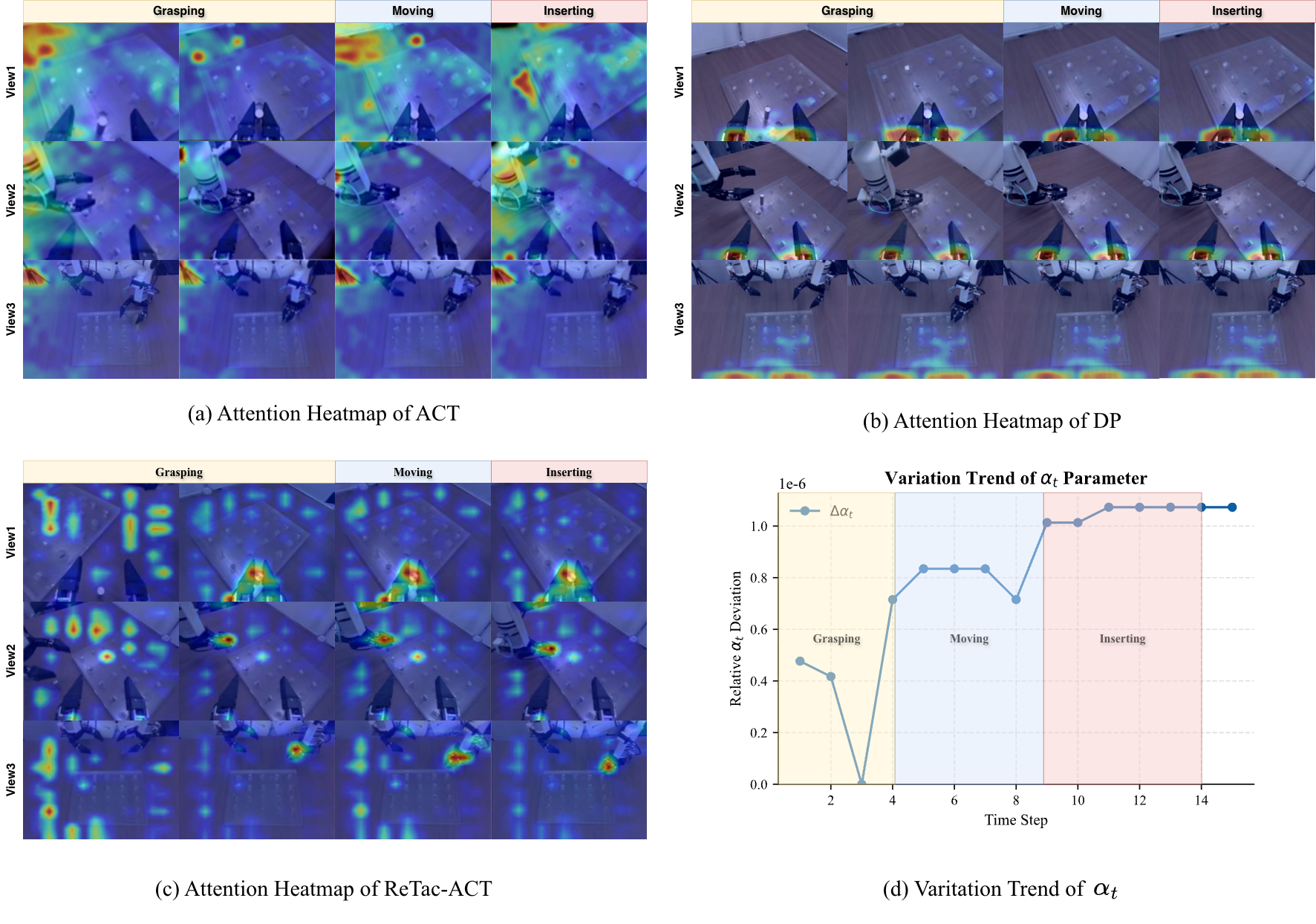}
  \caption{Visual Attention Heatmap Comparison and Parameter Evolution. (a) Attention heatmaps of ACT, showing attention scattered on background regions across three task stages. (b) Attention heatmaps of DP. (c) Attention heatmaps of our ReTac-ACT. (d) Evolution of $\alpha_t$ during the task.}
  \label{fig:attention_map}
\end{figure*}

\emph{1) Absolute Performance:}
At the tightest clearance (0.1\,mm), ReTac-ACT achieves 80\% insertion success while ACT drops to only 15\% and DP fails completely (0\%).
This 0.1\,mm clearance approaches the tolerance range of ISO IT6--IT7 grades, suggesting potential applicability to precision assembly scenarios.

\emph{2) Degradation Rate:}
As clearance tightens from 3\,mm to 0.1\,mm, DP degrades from 20\% to 0\%, ACT degrades from 40\% to 15\% (a 62.5\% relative drop), and ReTac-ACT degrades from 90\% to 80\% (only an 11\% relative drop).
This stark contrast in performance degradation highlights the indispensability of tactile sensing in precision assembly. As clearance tightens, pure-vision policies face exponentially growing challenges during the final insertion phase, which demands exceedingly stringent positional accuracy. Given the severe visual occlusion and the microscopic nature of the clearance at this critical stage, relying solely on visual information to guide and confirm the final assembly becomes practically impossible. Conversely, tactile sensing provides real-time contact feedback, enabling the policy to execute guided sub-millimeter corrections based on physical contact information rather than struggling with occluded and resolution-limited visual cues.

\emph{3) Grasp Reliability:}
ReTac-ACT maintains 0\% miss rate across all clearances, whereas ACT and DP consistently miss 40\% and 70\% of grasps respectively, regardless of clearance. This provides further evidence that our vision-tactile fusion benefits both grasping and insertion phases.

\subsection{Qualitative Analysis: Attention Visualization}
To understand how ReTac-ACT achieves superior performance, we visualize attention heatmaps during policy rollout (Fig.~\ref{fig:attention_map}).

\emph{1) State-Dependent Focus:}
While ACT exhibits scattered attention across background regions, ReTac-ACT consistently focuses on task-relevant objects (peg and hole).
This focused attention arises from our bidirectional cross-attention: tactile features query visual tokens to identify contact-relevant regions, while visual features reciprocally inform tactile interpretation.

\emph{2) Instant Modality Switching:}
A striking phenomenon occurs at the moment of contact ($t{=}T_{\text{contact}}$): the attention heatmap instantly concentrates and locks onto the contacted assembly components.
This phenomenon directly reflects the gains brought by integrating the tactile modality via vision-tactile cross-attention.
From free-space motion to the grasping position, the system receives no tactile contact information. However, the moment the object is contacted and tactile signals emerge, the tactile-to-visual cross-attention guides visual focus to spatially relevant regions, effectively suppressing background noise precisely when contact information becomes the critical information source. Simultaneously, observing the changes in the fusion weight parameter during the task, it is evident that as tactile information becomes available and the task state transitions from vision-guided to tactile-dominant, the proportion of the tactile fusion weight increases.

\section{Conclusion}

We presented ReTac-ACT, a vision-tactile imitation learning policy that addresses a fundamental challenge in precision assembly: the failure of vision-guided policies in contact-rich ``last-millimeter'' regions, where sub-millimeter corrections are critical but visual feedback is compromised by occlusion.
Our approach achieves superior precision assembly performance by integrating visual and tactile modalities through three synergistic innovations: bidirectional cross-attention for vision and tactile features enhancement, state-gated dynamic fusion for adaptive modality weighting, and tactile reconstruction for learning tactile representations.

Experiments on the real-world peg-in-hole assembly tasks demonstrate that ReTac-ACT achieves 90\% peg-in-hole success (outperforming ACT by a factor of 2.25, and DP/pi05 by a factor of 4.5) and maintains 80\% success at 0.1\,mm clearance where ACT drops to 15\%.
Comprehensive ablation studies validate that each architectural component is indispensable.

Limitations and Future Work:
Our current evaluation focuses on cylindrical pegs; extending to all five NIST ATB shapes remains future work.
Specifically, we plan to evaluate on non-axially-symmetric shapes (e.g., square, hexagonal pegs) where tactile geometry plays an even more critical role in orientation alignment.
We also plan to investigate real-to-sim and sim-to-real transfer of tactile representations and integration with large-scale VLA pre-training to combine generalist knowledge with specialized tactile reasoning.

\section*{Acknowledgments}
This work was supported by the Brain Science and Brain-like Intelligence Technology---National Science and Technology Major Project (Grant No.\ 2025ZD0215600), in part by the National Natural Science Foundation of China under Grant No.\ 62573063 and No.\ 62536001, and the Open Foundation of the State Key Laboratory of Precision Space-time Information Sensing Technology (No.\ STSL2025-B-07-01(C)).

\bibliographystyle{IEEEtran}
\bibliography{references}

@article{ref1,
  author    = {Intelligence, P. and others},
  title     = {$\pi_{0.5}$: A Vision-Language-Action Model with Open-World Generalization},
  journal   = {arXiv preprint arXiv:2504.16054},
  year      = {2025},
  doi       = {10.48550/arXiv.2504.16054}
}

@article{ref2,
  author    = {Black, K. and others},
  title     = {$\pi_0$: A Vision-Language-Action Flow Model for General Robot Control},
  journal   = {arXiv preprint arXiv:2410.24164},
  year      = {2026},
  doi       = {10.48550/arXiv.2410.24164}
}

@article{ref4,
  author    = {Kobayashi, T. and Kobayashi, M. and Buamanee, T. and Uranishi, Y.},
  title     = {Bi-LAT: Bilateral Control-Based Imitation Learning via Natural Language and Action Chunking with Transformers},
  journal   = {arXiv preprint arXiv:2504.01301},
  year      = {2025},
  doi       = {10.48550/arXiv.2504.01301}
}

@inproceedings{ref5,
  author    = {Yang, S. and others},
  title     = {BiTLA: A Bimanual Tactile-Language-Action Model for Contact-Rich Robotic Manipulation},
  booktitle = {Proceedings of the 1st International Workshop on Multi-Sensorial Media and Applications},
  publisher = {ACM},
  year      = {2025},
  pages     = {12--17},
  doi       = {10.1145/3728485.3759237}
}

@article{ref6,
  author    = {Li, Q. and others},
  title     = {CogACT: A Foundational Vision-Language-Action Model for Synergizing Cognition and Action in Robotic Manipulation},
  journal   = {arXiv preprint arXiv:2411.19650},
  year      = {2024},
  doi       = {10.48550/arXiv.2411.19650}
}

@article{ref7,
  author    = {Chi, C. and others},
  title     = {Diffusion Policy: Visuomotor Policy Learning via Action Diffusion},
  journal   = {arXiv preprint arXiv:2303.04137},
  year      = {2024},
  doi       = {10.48550/arXiv.2303.04137}
}

@article{ref8,
  author    = {Yuan, W. and Dong, S. and Adelson, E.},
  title     = {GelSight: High-Resolution Robot Tactile Sensors for Estimating Geometry and Force},
  journal   = {Sensors},
  volume    = {17},
  number    = {12},
  pages     = {2762},
  year      = {2017},
  doi       = {10.3390/s17122762}
}

@article{ref9,
  author    = {Zhao, T. Z. and Kumar, V. and Levine, S. and Finn, C.},
  title     = {Learning Fine-Grained Bimanual Manipulation with Low-Cost Hardware},
  journal   = {arXiv preprint arXiv:2304.13705},
  year      = {2023},
  doi       = {10.48550/arXiv.2304.13705}
}

@article{ref10,
  author    = {Levine, S. and Finn, C. and Darrell, T. and Abbeel, P.},
  title     = {End-to-End Training of Deep Visuomotor Policies},
  journal   = {arXiv preprint arXiv:1504.00702},
  year      = {2016},
  doi       = {10.48550/arXiv.1504.00702}
}

@inproceedings{ref11,
  author    = {Yu, S. and Kelvin, L. and Xiao, A. and Duan, J. and Soh, H.},
  title     = {Octopi: Object Property Reasoning with Large Tactile-Language Models},
  booktitle = {Robotics: Science and Systems XX},
  publisher = {Robotics: Science and Systems Foundation},
  year      = {2024},
  doi       = {10.15607/RSS.2024.XX.066}
}

@article{ref13,
  author    = {Black, K. and Galliker, M. Y. and Levine, S.},
  title     = {Real-Time Execution of Action Chunking Flow Policies},
  journal   = {arXiv preprint arXiv:2506.07339},
  year      = {2025},
  doi       = {10.48550/arXiv.2506.07339}
}

@article{ref14,
  author    = {Song, W. and others},
  title     = {ReconVLA: Reconstructive Vision-Language-Action Model as Effective Robot Perceiver},
  journal   = {arXiv preprint arXiv:2508.10333},
  year      = {2025},
  doi       = {10.48550/arXiv.2508.10333}
}

@article{ref15,
  author    = {Li, H. and others},
  title     = {See, Hear, and Feel: Smart Sensory Fusion for Robotic Manipulation},
  journal   = {arXiv preprint arXiv:2212.03858},
  year      = {2022},
  doi       = {10.48550/arXiv.2212.03858}
}

@article{ref16,
  author    = {Shukor, M. and others},
  title     = {SmolVLA: A Vision-Language-Action Model for Affordable and Efficient Robotics},
  journal   = {arXiv preprint arXiv:2506.01844},
  year      = {2025},
  doi       = {10.48550/arXiv.2506.01844}
}

@article{ref17,
  author    = {Huang, J. and Wang, S. and Lin, F. and Hu, Y. and Wen, C. and Gao, Y.},
  title     = {Tactile-VLA: Unlocking Vision-Language-Action Model’s Physical Knowledge for Tactile Generalization},
  journal   = {arXiv preprint arXiv:2507.09160},
  year      = {2025},
  doi       = {10.48550/arXiv.2507.09160}
}

@article{ref19,
  author    = {Si, Z. and Yuan, W.},
  title     = {Taxim: An Example-Based Simulation Model for GelSight Tactile Sensors},
  journal   = {arXiv preprint arXiv:2109.04027},
  year      = {2021},
  doi       = {10.48550/arXiv.2109.04027}
}

@article{ref20,
  author    = {Calandra, R. and others},
  title     = {The Feeling of Success: Does Touch Sensing Help Predict Grasp Outcomes?},
  journal   = {arXiv preprint arXiv:1710.05512},
  year      = {2025},
  doi       = {10.48550/arXiv.1710.05512}
}

@article{ref21,
  author    = {Hao, P. and others},
  title     = {TLA: Tactile-Language-Action Model for Contact-Rich Manipulation},
  journal   = {Robot Learning},
  pages     = {1--1},
  year      = {2026},
  doi       = {10.55092/rl20260001}
}

@article{ref22,
  author    = {Zhu, X. and Huang, B. and Li, Y.},
  title     = {Touch in the Wild: Learning Fine-Grained Manipulation with a Portable Visuo-Tactile Gripper},
  journal   = {arXiv preprint arXiv:2507.15062},
  year      = {2025},
  doi       = {10.48550/arXiv.2507.15062}
}

@article{ref23,
  author    = {George, A. and Gano, S. and Katragadda, P. and Farimani, A. B.},
  title     = {VITaL Pretraining: Visuo-Tactile Pretraining for Tactile and Non-Tactile Manipulation Policies},
  journal   = {arXiv preprint arXiv:2403.11898},
  year      = {2024},
  doi       = {10.48550/arXiv.2403.11898}
}

@article{ref24,
  author    = {Liu, F. and Li, C. and Qin, Y. and Xu, J. and Abbeel, P. and Chen, R.},
  title     = {ViTaMIn: Learning Contact-Rich Tasks through Robot-Free Visuo-Tactile Manipulation Interface},
  journal   = {arXiv preprint arXiv:2504.06156},
  year      = {2025},
  doi       = {10.48550/arXiv.2504.06156}
}

@article{ref25,
  author    = {Bi, J. and Ma, K. Y. and Hao, C. and Shou, M. Z. and Soh, H.},
  title     = {VLA-Touch: Enhancing Vision-Language-Action Models with Dual-Level Tactile Feedback},
  journal   = {arXiv preprint arXiv:2507.17294},
  year      = {2025},
  doi       = {10.48550/arXiv.2507.17294}
}

@article{ref26,
  author    = {Zhang, C. and Hao, P. and Cao, X. and Hao, X. and Cui, S. and Wang, S.},
  title     = {VTLA: Vision-Tactile-Language-Action Model with Preference Learning for Insertion Manipulation},
  journal   = {arXiv preprint arXiv:2505.09577},
  year      = {2025},
  doi       = {10.48550/arXiv.2505.09577}
}

@article{ref27,
  author    = {Johansson, R. S. and Flanagan, J. R.},
  title     = {Coding and Use of Tactile Signals from the Fingertips in Object Manipulation Tasks},
  journal   = {Nature Reviews Neuroscience},
  volume    = {10},
  number    = {5},
  pages     = {345--359},
  year      = {2009}
}

@article{ref28,
  author    = {Howe, R. D.},
  title     = {Tactile Sensing and Control of Robotic Manipulation},
  journal   = {Advanced Robotics},
  volume    = {8},
  number    = {3},
  pages     = {245--261},
  year      = {1993},
  doi       = {10.1163/156855394X00356}
}

@inproceedings{ref29,
  author    = {Bicchi, A. and Bergamasco, M. and Dario, P. and Fiorillo, A.},
  title     = {Integrated Tactile Sensing for Gripper Fingers},
  booktitle = {Proceedings of the International Conference on Robot Vision and Sensory Control},
  year      = {1988}
}

@article{ref30,
  author    = {Romano, J. M. and Hsiao, K. and Niemeyer, G. and Chitta, S. and Kuchenbecker, K. J.},
  title     = {Human-Inspired Robotic Grasp Control with Tactile Sensing},
  journal   = {IEEE Transactions on Robotics},
  volume    = {27},
  number    = {6},
  pages     = {1067--1079},
  year      = {2011},
  doi       = {10.1109/TRO.2011.2162271}
}

@inproceedings{ref31,
  author    = {Levine, S. and Wagener, N. and Abbeel, P.},
  title     = {Learning Contact-Rich Manipulation Skills with Guided Policy Search},
  booktitle = {2015 IEEE International Conference on Robotics and Automation (ICRA)},
  publisher = {IEEE},
  year      = {2015},
  pages     = {156--163},
  doi       = {10.1109/ICRA.2015.7138994}
}

@article{ref32,
  author    = {Lee, M. A. and others},
  title     = {Making Sense of Vision and Touch: Self-Supervised Learning of Multimodal Representations for Contact-Rich Tasks},
  journal   = {arXiv preprint arXiv:1810.10191},
  year      = {2019},
  doi       = {10.48550/arXiv.1810.10191}
}

@misc{manipulationnet2025,
  author       = {Kaiyu Hang and Kenneth Kimble and others},
  title        = {{ManipulationNet}: Benchmarking Robotic Manipulation in the Real World at Scale with Any Robot at Any Time and Anywhere},
  howpublished = {[Online]. Available: \url{https://manipulation-net.org}},
  month        = mar,
  year         = {2026},
  note         = {Accessed: Mar. 2026}
}



\end{document}